# Short-Term Forecasting COVID-19 Cases In Turkey Using Long Short-Term Memory Network

Selahattin Serdar HELLİ, Çağkan DEMİRCİ, Onur ÇOBAN, Andaç HAMAMCI
Department of Biomedical Engineering,
Yeditepe University,
Istanbul, Turkey
selahattinserdar.helli@std.yeditepe.edu.tr , cagkan.demirci@std.yeditepe.edu.tr,
onur.coban@std.yeditepe.edu.tr, andac.hamamci@yeditepe.edu.tr

*Abstract—* COVID-19 has been one of the most severe diseases, causing a harsh pandemic all over the world, since December 2019. The aim of this study is to evaluate the value of Long Short-Term Memory (LSTM) Networks in forecasting the total number of COVID-19 cases in Turkey. The COVID-19 data for 30 days, between March 24 and April 23, 2020, are used to estimate the next fifteen days. The mean absolute error of the LSTM Network for 15 days estimation is 1,69±1.35%. Whereas, for the same data, the error of the Box-Jenkins method is 3.24±1.56%, Prophet method is 6.88±4.96% and Holt-Winters Additive method with Damped Trend is 0.47±0.28%. Additionally, when the number of deaths data is also provided with the number of total cases to the input of LSTM Network, the mean error reduces to 0.99±0.51%. Consequently, addition of the number of deaths data to the input, results a lower error in forecasting, compared to using only the number of total cases as the input. However, Holt-Winters Additive method with Damped Trend gives superior results to LSTM Networks in forecasting the total number of COVID-19 cases.

## I. Introduction

COVID-19 is an infectious disease caused by severe acute respiratory syndrome coronavirus 2 (SARS-CoV-2). It was first identified in December 2019 in Wuhan, China, and has since spread globally, resulting in an ongoing pandemic. The governments have been making an effort to oversee the cases and fatalities in their countries; to forecast their future position, evaluate how effective taken decisions are, what would happen if the public conforms to the restrictions. The estimation of COVID-19 cases has been the subject of many studies since the novel coronavirus disease.

Time series forecasting methods have been applied in a variety of fields such as finance, supply and demand prediction, signal processing, tracking corporate business metrics, and health monitoring. As a given vector of historical time series values $T = [t_1, t_2, ..., t_N] \in R^N$, a prevalent task in time series analysis is to forecast future values $t'_{N+1}, t'_{N+2}, ...$ based on the time-series data. Time series forecasting methods basically can be defined under two main categories: traditional statistical methods and methods based on machine learning models [6].

A variety of statistical and machine learning models, such as Box-Jenkins (ARIMA), Prophet, Holt-Winters Additive Model (HWAAS), and Gauss Model, have been used to forecast the number of cases of COVID-19 [1,2,5,18].

Hu, et.al. developed a modified stacked auto-encoder for modeling the transmission dynamics of the epidemics. They applied this model to real-time forecasting the curves of cumulative confirmed cases of Covid-19 across China from January 20 to April 20, 2020. Using the multiple-step forecasting, the estimated average errors of 6-step, 7-step, 8-step, 9-step and 10-step forecasting were 1.64%, 2.27%, 2.14%, 2.08%, 0.73%, respectively [2].

In Yonar, et.al. [1], the number of COVID-19 cases of the selected G8 countries, Germany, United Kingdom, France, Italy, Russian, Canada, Japan, and Turkey between January 22 and March 22, 2020 have been estimated using curve estimation models, namely; Box-Jenkins (ARIMA) and Brown/Holt linear exponential smoothing. The mean absolute percentage error of ARIMA (1,4,0) for Turkey was reported as % 2.578 [1].

According to Papastefanopoulos, et.al., Holt-Winters Additive Model (HWAAS) model showed the best performance compared to the others models in forecasting the active cases per population in Turkey [18].

In the last years, despite many models, Long Short-Term Memory (LSTM) neural network has shown great performance in time series forecasting [4,7]. The LSTM neural network is a special kind of Recurrent Neural Network (RNN), capable of learning long-term dependencies. LSTMs were introduced by Hochreiter&Schmidhuber [9] and were refined and popularized by many studies. The LSTM architecture was able to take care of the vanishing gradient problem in the traditional RNN.

The main purpose of this study is to help nations and their health care systems by forecasting positive cases in Turkey using LSTM neural network. We propose that LSTM networks are effective tools in short-term time series forecast the COVID-19 confirmed cases in Turkey.

## II. METHOD

### A. Dataset

The official total number of confirmed COVID-19 cases and deaths between 11 March 2020 and 8 May 2020, which are reported by the Republic of Turkey Ministry of Health, are used in the study [8]. The links to dataset, as well as the source codes are provided in the Appendix.

### B. The LSTM Architecture

The network consists of an LSTM Block which is followed by the dense layer (the regular fully-connected neural network layer). The LSTM layer has 32 units which is the dimensionality of the hidden state at the proposed network. The network is designed as a cycle for recursive multi-step time series forecasting values of multiple future days [4]. The diagram of the network architecture is illustrated in Fig.1. In this cycle, the output is to be used as a new input to complete multi-step forecasting. Finally, all outputs, before going into the loop, are stored as results. The input is normalized to the range of 0 to 1. The optimization is done with the "Adam" [19]. The network is trained with 2000 epochs and one batch size.

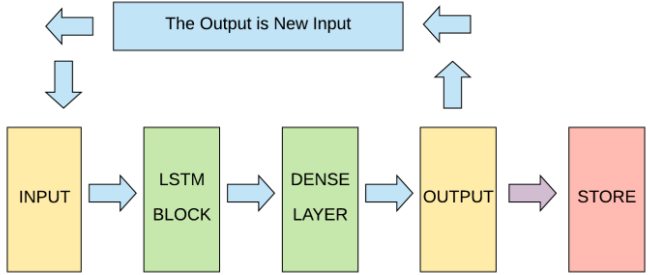

Fig. 1. The Network Architecture for the Recursive Forecasting

The LSTM block, given in the diagram in Fig. 2., has four main parts: an input gate, a forget gate, an output gate, and self-connected memory cells. The input gate decides which new information will enter the memory cell. The output gate decides what cell activations to filter at the output. The forget gate helps the network to forget the past input data and reset the memory cell [4]. The LSTM block, used in this study, is defined by the following equations [3]:

$$i_t = \sigma(W_{ix}x_t + W_{ih}h_{t-1} + b_i)$$
$$f_t = \sigma(W_{fx}x_t + W_{fh}h_{t-1} + b_f)$$
$$o_t = \sigma(W_{ox}x_t + W_{oh}h_{t-1} + b_o) \quad (1)$$
$$c_t = f_t \odot c_{t-1} + i_t \odot g(W_{cx}x_t + W_{ch}h_{t-1} + b_c)$$
$$h_t = o_t \odot g(c_t)$$

where $i_t$, $f_t$, $o_t$, and $c_t$ represent input gate, forget gate, output gate, and memory cell, respectively. Input, and hidden layer activation at time $t$ are $x_t$ and $h_t$. $W$'s and $b$'s denote weights and biases, respectively. $\sigma$ represents the logistic sigmoid function. In this study, exponential linear unit (elu) is used as the $g$ function. $\odot$ denotes element-wise multiplication.

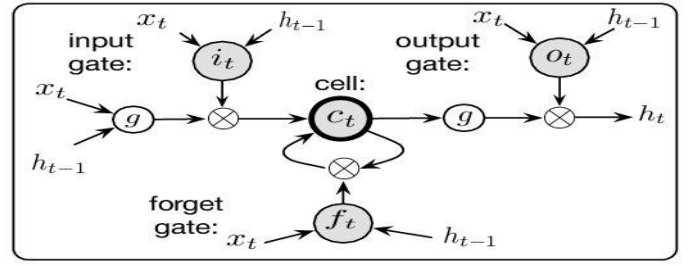

Fig. 2. The diagram of LSTM Block.

### C. Forecasting Schemes of The LSTM Neural Network

Three different forecasting schemes have been used:

*1) The First Schema (u1):* The first schema is one-step prediction of the next day. The total number of cases in Turkey is used to estimate the next day's result. The input data is based on 30 days between March 24 and May 8, 2020.

*2) The Second Schema (u2):* The total number of cases for the next fifteen days are estimated using recursive approach in the second schema (see Fig. 1.). In the u2 schema, the number of total cases for 30 days between March 24 and April 23, 2020, is used as the input data.

*3) The Third Schema (u3):* The third schema is to estimate next 15 days recursively, similar to u2. The difference is that the total number of deaths was added to the input data of the third schema. The total number of confirmed cases and deaths for 30 days, between March 24 and April 23, 2020 in Turkey are used as the input data.

### D. Other Models

In order to compare with the proposed LSTM neural network, the Box-Jenkins (ARIMA) method, Prophet, Holt-Winter's Additive Model with Damped Trend are implemented. In these models, the total number of cases for 30 days in Turkey, between March 24 and April 23, 2020, are used to estimate the next fifteen days.

*1) Box-Jenkins Method (ARIMA):* The Autoregressive Integrated Moving Average (ARIMA) model is a combination of the differenced autoregressive model with the moving average model. The ARIMA model is effective in fitting historical data with this combination approach and helpful for forecasting future points in a time series [11]. Box-Jenkins method contains ARIMA models that were applied to the series that are non-stationary but are made stationary with the difference operation on the series, and widely used for time series analysis [2,10]. In this study, ARIMA model is used with (6,1,0) orders.

*2) Prophet, Automatic Forecasting Procedure:* Prophet, Automatic Forecasting Procedure [13] is a method of forecasting time series data based on an additional model where non-linear trends are fit with yearly, weekly, daily, seasonal, and, holiday effects. Also, it shows the best performance with time series that have strong seasonal effects and historical data of several seasons[12]. In its structure, it is a regression procedure with interpretable parameters. Prophet, Automatic

Forecasting Procedure is robust to lacking data and shifts in the trend, and typically performs outliers well [13].

*3) Holt-Winter's Additive Model With Damped Trend*: Holt (1957) and Winters (1960) [16] extended Holt's method [15] to capture seasonality. Also, Gardner & McKenzie (1985) came up with a parameter that "dampens" the trend to a flat line sometime in the future [14]. A method that often provides accurate forecasts for seasonal data is the Holt-Winters method with a damped trend and multiplicative seasonality [16,17]. In this study 0.96 is used as the ratio of damping slope.

### III. RESULTS

The whole training process of all schemes of LSTMs has taken about 15 minutes with an NVidia GeForce GTX 960M GPU, whereas the other models were completed in a short duration.

For each method, the error in estimated total number of cases are quantified by the absolute percentage error (APE) metric, according to (3) and given in Table 1.

$$APE = \left|\frac{Actual - Forecast}{Actual}\right| \times 100 \quad (3)$$

In order to compare the effect of the chosen activation function g in (1), the LSTM schemes are implemented with hyperbolic tangent function and exponential linear unit (elu). Mean estimation errors are given in Table 1. In line with these results, for the rest of the study, the exponential linear unit (elu) is used as the activation function of the LSTM network.

TABLE I. MEAN ESTIMATION ERRORS OF LSTM SCHEMES IMPLEMENTED USING TWO DIFFERENT LSTM ACTIVATION FUNCTIONS

|  | U1 | U2 | U3 |
|---|---|---|---|
| Hyperbolic Tangent (TANH) | 0.81±0.51% | 3.33±2.76% | 3.71±2.89% |
| Exponential Linear Unit (ELU) | 0,70±0.30% | 1,69±1.35% | 0.99±0.51% |

TABLE II. THE ESTIMATION ERRORS OF THE METHODS

|  | The LSTM Network Schemes | | | Other Models | | |
|---|---|---|---|---|---|---|
|  | U1 | U2 | U3 | ARIMA (6,1,0) | PROPHET | HWAAS |
| 1. Day | 0.55 % | 0.45 % | 0.69 % | 0.52 % | 4.34 % | 0.12 % |
| 2. Day | 0.24 % | 0.40 % | 0.54 % | 1.42 % | 3.12 % | 0.11 % |
| 3. Day | 0.53 % | 0.88 % | 0.94 % | 1.81 % | 1.51 % | 0.25 % |
| 4. Day | 0.72 % | 1.46 % | 1.31 % | 2.03 % | 0.23 % | 0.71 % |
| 5. Day | 0.29 % | 1.36 % | 1.33 % | 2.43 % | 1.67 % | 0.82 % |
| 6. Day | 1.07 % | 0.54 % | 0.99 % | 2.64 % | 2.57 % | 0.38 % |
| 7. Day | 1.15 % | 0.30 % | 0.20 % | 2.63 % | 3.71 % | 0.14 % |
| 8. Day | 0.86 % | 0.58 % | 0.30 % | 2.56 % | 5.17 % | 0.19 % |
| 9. Day | 1.16 % | 1.23 % | 0.21 % | 2.90 % | 6.75 % | 0.33 % |
| 10. Day | 0.66 % | 1.39 % | 0.77 % | 3.85 % | 8.55 % | 0.64 % |
| 11. Day | 0.26 % | 2.08 % | 0.66 % | 4.65 % | 10.35 % | 0.93 % |
| 12. Day | 0.48 % | 2.57 % | 0.50 % | 4.88 % | 11.91 % | 0.97 % |
| 13. Day | 0.86 % | 3.02 % | 1.58 % | 4.86 % | 13.08 % | 0.64 % |
| 14. Day | 1.03 % | 4.36 % | 2.22 % | 5.41 % | 14.43 % | 0.47 % |
| 15. Day | 0.64 % | 4.67 % | 2.68 % | 6.06 % | 15.86 % | 0.34 % |
| MAPE[a] | 0.70±0.30 | 1.69±1.35 | 0.99±0.51 | 3.24±1.56 | 6.88±4.96 | 0.47±0.28 |

[a.] Mean Absolute Percentage Error (Mean±Std.Dev.)

The estimation results of different forecasting schemes of the LSTM network are plotted on the graph in Fig. 3. The estimation results of different methods are given in Table 2 and Fig. 4. In Fig. 4, U2 schema of the LSTM network is reported to have exactly the same input for all methods.

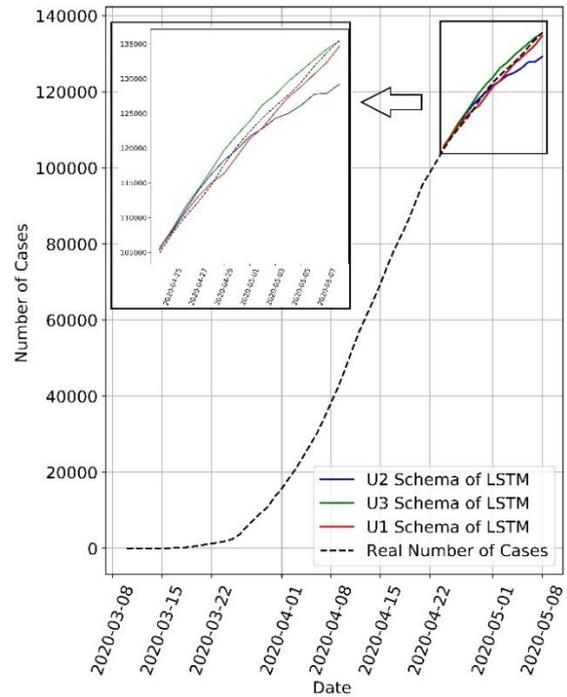

Fig. 3. The estimation results of different forecasting schemes used for LSTM Network with the real COVID-19 cases. U1 is the next-single day estimation, U2 is the recursive estimation of the next 15 days and U3 is the result of recursive estimation including also the number of deaths data as input.

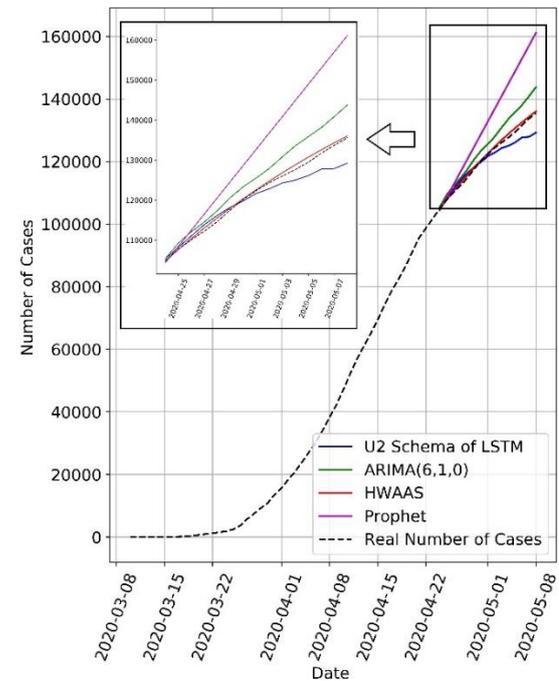

Fig. 4. The estimation results of different methods with the real COVID-19 cases, The models are: U2 Schema of LSTM, ARIMA (6,1,0), HWAAS with Damped Trend and Prophet.

## IV. DISCUSSION

In this study, the value of Long Short-Term Memory (LSTM) Networks in forecasting the total number of COVID-19 cases in Turkey was evaluated.

The LSTM Network was implemented with two different activation functions. Using exponential linear unit (elu) instead of common choice of hyperbolic tangent (tanh) resulted better performance in forecasting the total number of COVID-19 cases.

LSTMs neural network has shown more successful performance, compared to the ARIMA model (6,1,0) and Prophet. However, Holt-Winters Additive method with Damped Trend gave superior results to LSTM Networks in forecasting the total number of COVID-19 cases. Prophet has shown the worst performance.

Addition of the number of deaths data to the input of the LSTM Network, resulted a lower error in forecasting, compared to using only the number of total cases as the input.

In conclusion, it has been observed that the artificial intelligence neural network can be used in the medical field, especially for the prediction of conditions in a severe pandemic such as Covid-19. Therefore, nations can be more prepared to take the necessary precautions in advance, impose the required restrictions in time and be able to observe the possible scenarios that may occur in a case of alteration.

## ACKNOWLEDGMENTS

The authors thank Doğuş Öztok for editing the text.

## APPENDIX

All datasets as well as the necessary code used in this study is available at: https://github.com/SerdarHelli/SHORT-TERM-FORECASTING-COVID-19-IN-TURKEY